\documentclass{ecai}
\usepackage{times}
\usepackage{graphicx}
\usepackage{latexsym}
\usepackage{tabularx}

\ecaisubmission 

\usepackage{multicol}
\usepackage{amsmath,amssymb}
\usepackage{multirow}
\usepackage{graphicx}
\usepackage{algorithm}
\usepackage{algorithmic}
\usepackage{color,soul}
\usepackage{booktabs} 
\usepackage{tabularx}
\usepackage{subcaption}
\usepackage{mwe}
\usepackage{commath}
\usepackage{verbatim}
\usepackage{mathtools}
\usepackage{cuted}
\usepackage{flushend}
\usepackage{arydshln}
\usepackage{multirow}

\usepackage{cite}
\usepackage{amsmath,amssymb,amsfonts}
\usepackage{algorithmic}

\usepackage{textcomp}
\usepackage{xcolor}
\usepackage{authblk}
\usepackage{times}  
\usepackage{helvet}  
\usepackage{courier}  
\usepackage{url}  
\usepackage{graphicx}  

\usepackage{mathtools}
\usepackage{amsmath,amssymb,amsfonts}

\usepackage{caption}
\usepackage{amsthm,dsfont}
\newcommand{\ignore}[1]{}
\usepackage{color}
\usepackage{authblk}
\usepackage{graphicx}
\usepackage{float} 
\usepackage{multirow}
\usepackage{amsfonts}
\usepackage{comment}
\usepackage{fancyhdr}
\usepackage{rotating}

\def\BibTeX{{\rm B\kern-.05em{\sc i\kern-.025em b}\kern-.08em
    T\kern-.1667em\lower.7ex\hbox{E}\kern-.125emX}}

\begin{document}

\title{Neural Networks in Evolutionary Dynamic Constrained Optimization: Computational Cost and Benefits}

\author{Maryam Hasani-Shoreh\institute{University of Adelaide, email: maryam.hasanishoreh@adelaide.edu.au} \and Renato Hermoza Aragonés\institute{University of Adelaide, email: renato.hermozaargones@adelaide.edu.au} \and Frank Neumann\institute{University of Adelaide, email: frank.neumann@adelaide.edu.au} }

\maketitle

\begin{abstract}

Neural networks (NN) have been recently applied together with evolutionary algorithms (EAs) to solve dynamic optimization problems.
The applied NN estimates the position of the next optimum based on the previous time best solutions. After detecting a change, the predicted solution can be employed to move the EA’s population to a promising region of the solution space in order to accelerate convergence and improve accuracy in tracking the optimum.
While previous works show improvement of the results, they neglect the overhead created by NN.
In this work, we reflect the time spent for training NN in the optimization time and compare the results with a baseline EA.
We explore if by considering the generated overhead, NN is still able to improve the results, and under which conditions is able to do so. 

The main difficulties to train the NN are: 1) to get enough samples to generalize predictions for new data, and 2) to obtain reliable samples.
As NN needs to collect data at each time step, if the time horizon is short, we will not be able to collect enough samples to train the NN.
To alleviate this, we propose to consider more individuals on each time to speed up sample collection in shorter time steps.
In environments with high frequency of changes, the solutions produced by EA are likely to be far from the real optimum.
Using unreliable train data for the NN will, in consequence, produce unreliable predictions.
Also, as the time spent for NN stays fixed regardless of the frequency, a higher frequency of change will mean a higher produced overhead by the NN in proportion to the EA.
In general, after considering the generated overhead, we conclude that NN is not suitable in environments with high frequency of changes and/or short time horizons. However, it can be promising for the low frequency of changes, and especially for the environments that changes have a pattern.
\end{abstract}

\section{INTRODUCTION}
\label{sec:intro}
In this section, the background on the topic and our contribution are presented.
\subsection{Background}
Many real-world problems have uncertainties due to factors such as variation in the demand market, unpredicted events, variable resources, or estimated parameters that may change over time~\cite{branke2003designing}. These problems in which the objective function or/and the constraints change over time, are called as dynamic constrained optimization problems (DCOPs)~\cite{liu2008adaptive}.  
The goal is to find and track the optimum in each instance of the dynamic problem given a limited computational budget. One approach is to apply an independent optimization method to solve each problem instance separately, however, a more efficient approach solves them in a dynamic manner, in which the algorithm detects and responds to the changes on-the-fly~\cite{Nguyen20121}. 
Mathematically, the objective is to find a solution vector ($\vec{x} \in \mathbb{R}^D $) at each time period $t$ such that: $\min_{\vec{x}\in F_t} f(\vec{x}, t)$, where $f:S \rightarrow \mathbb{R}$ is a single objective function, and $t \in N^+$ is the current time period.  
$F_{t}=\{ \vec{x} \mid \vec{x} \in [L,U], g_i (\vec{x},t) \le 0$\}  is the feasible region at time $t$, where $L$ and $U$ are the boundaries of the search space and $g_i(x, t)$ is the linear $i$th inequality constraint at time $t$.
To tackle these problems evolutionary algorithms (EAs) are commonly used~\cite{Nguyen20121}. 
However, in order to apply previously proposed EAs in static domains for such dynamic problems, some adaptations are needed for them to handle dynamic environments. Mechanisms like change detection and the ability to react to the changes should be applied, otherwise the whole population may converge and stuck in an area of the search space without noticing the change. Previously proposed approaches include introducing~\cite{Goh_2009} or maintaining diversity~\cite{Bui2005}, memory-based approaches~\cite{Richter2013}, multi-population approaches~\cite{branke2000multi} or prediction methods~\cite{Bu_2016}. Previous work on prediction has used different methods including Markov chains~\cite{markov2008evolutionary}, Kalman filters~\cite{kalman2008tracking}, linear~\cite{autoreg2006dynamic} and nonlinear regression techniques~\cite{nonlinearreg2009improving}, and recently neural networks (NN) become increasingly popular~\cite{liu2019neural,meier2018prediction,jiang2017transfer,meier2019uncertaint}.
These works have been applied in a variety of optimization classes including multi-objective optimization~\cite{ahrari2019new,zhou2013population}, discrete optimization\cite{simoes2014prediction}, dynamic constrained optimization~\cite{Bu_2016}, and time-linkage problems~\cite{bosman2007learning}.
 
All of the previous works show how environmental change pattern can be extracted from the previous environments to provide effective guidance for the EA to predict the future optimum.
For instance, in~\cite{kalman2008tracking} the Kalman filter is adopted to model the movement of the optimum and predict the possible optimum in new environments. Similarly, in~\cite{markov2008evolutionary} linear regression is used to estimate the time of the next change and Markov chains is adopted to predict new optimum based on the previous times optimum. Likewise, in~\cite{zhou2013population} the center points of Pareto sets in past environments are used as data to simulate the change pattern of the center points by using a regression model. In other works~\cite{liu2019neural,jiang2017transfer}, where the change pattern is not stable, it is proposed to directly construct a transfer model of the solutions/fitness, considering the correlation and difference between the two consecutive environments. 


\subsection{Our contribution}
What is neglected in previous works is the time used for training and calling the predictor.
In one recent work~\cite{liu2018neural}, the time spent for training the NN is reported, however, it is not compared to the overall optimization time. Such a comparison is needed, to reflect the overhead caused by using NN.
In the relevant literature of dynamic problems, often a change is designed to happen after a constant number of fitness evaluations or generations~\cite{Nguyen20121}. But we need to consider the difference between the algorithm using NN and the baseline algorithm in terms of the real computational cost.
In some real-world problems~\cite{branke2003designing}, the condition that leads to the dynamic behaviour of the problem, happens after a time constraint (for instance prices are updated hourly in a power market). In this situation, we want an optimization algorithm to achieve an optimum solution in a limited time budget, regardless of the number of fitness evaluations.
In particular, time is important to be accounted when using a NN, since by including several stages (data collection, training and predicting new solutions) can produce a noticeable time overhead in the optimization.
Therefore, we propose to create a change after an actual running time. With this, the time spent for training NN, is subtracted from the EA time. In consequence, all the methods have the same time budget for overall optimization in each time. The purpose is to observe, considering the assigned time to NN that is indeed taken from the EA time for optimization, if still NN helps the EA to improve the results.

Aside from the time constraint, our other concerns are regarding collecting sufficient samples to generalize predictions for new data, and the reliability of the samples.
For those dynamic problems that the overall time horizon is short, we are not able to collect enough samples to train the NN in proper time.
To alleviate this, we propose to consider more individuals on each time to speed up sample collection in shorter time steps.
In problems with high frequency of changes, the solutions produced by EA at the end of each time are likely to be far from the real optimum.
In such cases, using unreliable train data for the NN, in consequence, will produce unreliable predictions.
Also, as the time spent for NN stays fixed regardless of the frequency, a higher frequency will mean a higher produced overhead by the NN in proportion to the EA.

We choose differential evolution (DE) as our baseline algorithm as it has shown competitive results in constrained and dynamic optimization~\cite{Ameca-AlducinHB18}. 
Using this baseline, we experiment with different NN specifications.
We explore how to introduce predicted solutions to population and the effect of the number of individuals introduced to be replaced on each change.

The remainder of the paper is as follows. Section~\ref{sec:Prim} introduces DE algorithm and the NN design. Experimental setup will be presented in Section~\ref{sec:ExperimentalSetup}. Experimental results are reviewed in Section~\ref{sec:resultsAnalysis} and finally in Section~\ref{sec:conc} the results are concluded.

\section{PRELIMINARIES}
\label{sec:Prim}
In this section, an overview of the adapted differential evolution (DE) algorithm to solve DCOPs and the design of the NN are presented.

\subsection{Differential evolution for dynamic problems}
\label{subsec:DE}
Differential evolution (DE) is a stochastic search algorithm that is simple, reliable and fast which showed competitive results in constrained and dynamic optimization~\cite{Ameca-AlducinHB18}. Each vector $\vec{x}_{i, G}$ in the current population (called as target vector at the moment of the reproduction) generates one trial vector $\vec{u}_{i, G}$ by using a mutant vector $\vec{v}_{i,G}$. The mutant vector is created applying $\vec{v}_{i,G}= \vec{x}_{r0,G} + F (\vec{x}_{r1,G} - \vec{x}_{r2,G})$,
where $\vec{x}_{r0,G}$, $\vec{x}_{r1,G}$, and $\vec{x}_{r2,G}$ are vectors chosen at random from the current population ($r0 \neq r1 \neq r2 \neq i$); $\vec{x}_{r0,G}$ is known as the base vector and $\vec{x}_{r1,G}$, and $\vec{x}_{r2,G}$ are the difference vectors and $F>0$ is a parameter called scale factor. The trial vector is created by the recombination of the target vector and mutant vector using a crossover probability $CR \in [0,1]$. 
In this paper, a simple version of DE called DE/rand/1/bin variant is chosen; where ``rand" indicates how the base vector is chosen, ``1" represents  how many vector pairs will contribute in differential mutation, and ``bin" is the type of crossover (binomial in our case).
Feasibility rules~\cite{deb2000efficient} is applied for the constraint handling. 

In addition to constraint handling, the algorithms in DCOPs need a mechanism to detect the changes.
In the literature, re-evaluation of the solutions is the most common change-detection approach~\cite{Nguyen20121}. The algorithm regularly re-evaluates specific solutions (in this work the first and the middle individual of the population) to detect changes in their function values or/and the constraints. 
If a change is detected, then the change reaction approach will be activated. In this work, two approaches are considered. In the first approach, called noNN, the whole population is re-evaluated. In the second approach (detail explanations in next Section), 
some individuals of the population will be replaced with the predicted solutions and the rest of the individuals are re-evaluated.

\subsection{Neural network design}
\label{subsec:NN}
\begin{figure}[t!]
    \centering
    \begin{subfigure}[t]{0.5\textwidth}
        \centering
        \includegraphics[width=1.45in]{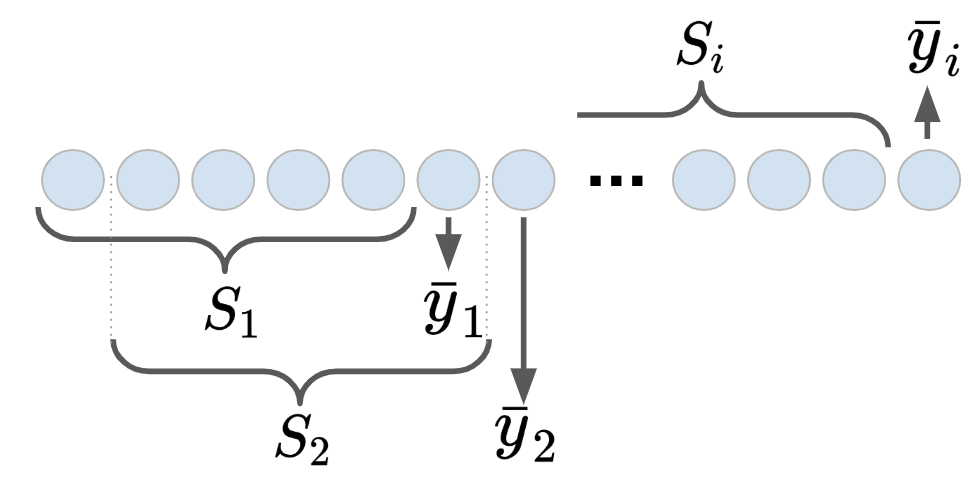}
        \caption{$n_t$ of previous times are used to predict next optimum}
        \label{fig:trainwineffect}
    \end{subfigure}
    \begin{subfigure}[t]{0.5\textwidth}
        \centering
        \includegraphics[height=0.7in]{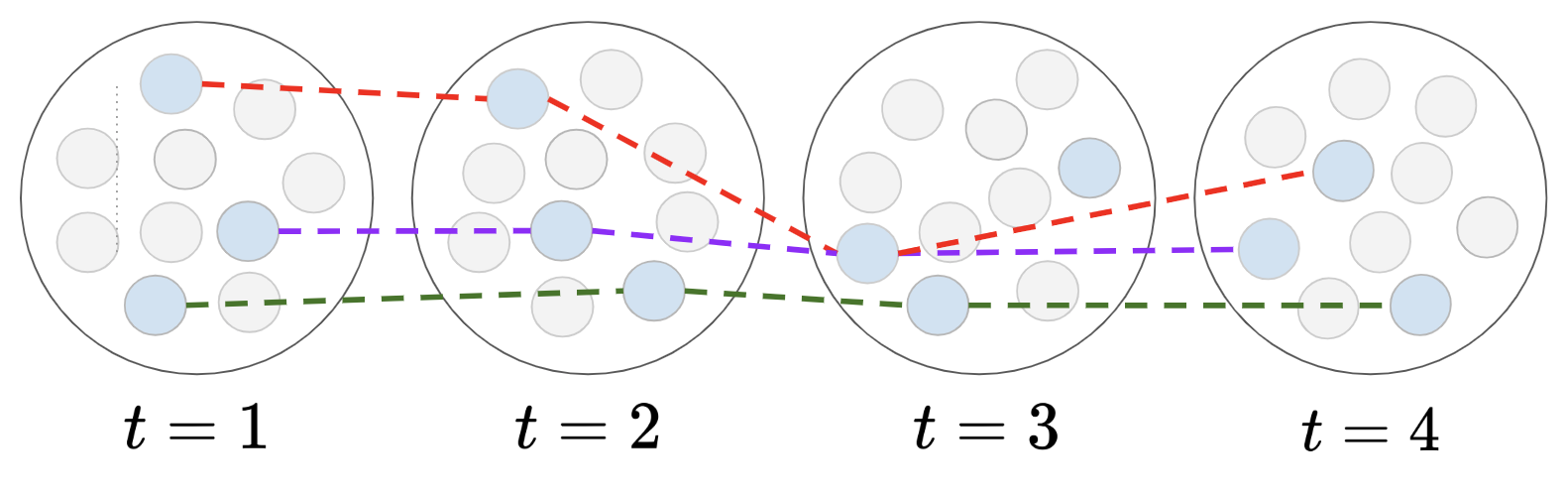}
        \caption{$k$-best individuals of each time are selected to train NN}
        \label{fig:samplesizeeffect}
    \end{subfigure}
    \caption{Building samples for NN}
\end{figure}

NN is intended to precisely model the optimum movement to make a reliable forecast of the future optimum position. To do so, the best solutions of the previous change periods found by the EA are required to build a time series $(\vec{x_0},...,\vec{x_{t−2}},\vec{x_{t−1})}$ for which the optimum $\vec{x_t}$ of the next change period $t$ has to be predicted (Figure~\ref{fig:trainwineffect}). To learn the change pattern of the optimum position, NN will go through a training process.
To train the network, $k$-best individuals (Figure~\ref{fig:samplesizeeffect}: example with $k=3$) of each time are collected for a couple of the previous times (based on a time-window ($n_t$)). 
There is the question of how far into the past should information be used to base the prediction upon.
In~\cite{liu2018neural}, the results of changes in $n_t$ show the addition of the older data brings noise
and misleads the NNs. It is concluded that the accumulation of old data is useful only to extract the overall environment change information. Therefore, a suggestion is, when constructing the training set, to select the data that has strong correlation to the predicted targets. 
In this work, $n_t=5$ is chosen for the experiments. 
For future work, we plan to explore the effect of time window ($n_t$). Considering an effective time window in which the shape of changes has a pattern is effective, as in some real-world problems the form of the dynamism could change overtime.
In addition, we plan to apply relational NN that considers priority for the data based on the distance to the predicted value in a time-series prediction (higher priority for closer ones accordingly).

We consider two cases: first one collects only one best individual ($k=1$) for 5 previous times ($n_t=5$) and predict the next one (Figure~\ref{fig:trainwineffect}). The second procedure considers $k$-best individuals of each time for $n_t=5$ and considers a combination of all possibilities ($k^{n_t}$) to build training data (Figure~\ref{fig:samplesizeeffect}). In the latter case, the samples are collected in faster speed. However, we consider to limit the number of samples collected by choosing a random subset of the above mentioned combination. As if we do not consider limits, the time spent for training data exponentially increases due to the large number of samples collected.
Also, as we have enough number of samples when using $k>1$, we can limit the NN to use the samples from $n_w$ previous changes. 
However, for the case with $k=1$, we keep collecting data and thus do not consider limits for number of collected samples ($n_w=\infty$) as otherwise the amount of samples will remain too low. 
Notice that, for the first environment changes we have a small amount of samples.
Hence, it is difficult for the NN to generalize from these data.
To avoid this, we wait until a minimum amount of samples are collected in order to train the NN.
We call this min\_batch size and empirically assign it a value of 20.
When $k=1$ we have to wait a large number of time changes in order to start using the NN.
But for the case where $k > 1$, we collect samples in faster speed, so the time lag to start using the NN is shorter.

The structure of the applied neural network has two hidden layers.
The first layer takes as an input an individual position $\vec{x_i}$ with $d$ dimensions and outputs a hidden representation $h_i$ of the individual with 4 dimensions.
As the network uses the last 5 times best individuals to predict a next one (Figure~\ref{fig:trainwineffect}), the first layer is applied to each of these 5 individuals ${\vec{x_1},...,\vec{x_5}}$ independently.
As a result, we obtain 5 hidden representation with 4 dimensions ${h_1,...,h_5}$; to aggregate their information, we choose to concatenate them into a variable $H$ with $4 \times 5$ dimensions. 
The second layer takes $H$ as input and then outputs a prediction with $d$ dimensions, representing the next best individual.
The layer one has rectified linear units (ReLU) activation function and the second layer has a linear output without activation function.
To train the network, we use mean squared error as a loss function. 
The predicted solution or its neighboring positions then can be used by EA to intensify the search in that region of the solution space. The mechanism to insert the predicted solutions in population can be either by replacing the worst individuals of the population, or random individuals.

\section{EXPERIMENTAL SETUP}
\label{sec:ExperimentalSetup}
In this section, designed experiments, the applied performance indicators, the test problems and the applied parameters are reviewed.

\subsection{Designed experiments}
Regarding to integration of NN with DE algorithm, there are a couple of experiments designed as follows.
\begin{itemize}
    \item \textbf{Frequency changes:} 
    In this experiment, we observe how the frequency of changes will affect the results. The frequency of change, denoted by $\tau$, represents the width that each time lasts. Notice that when we refer to higher frequencies of change, we mean lower values for $\tau$, since higher frequencies of change happens when there is shorter time interval between each change ($\tau$).
     Three frequencies of change: 0.5, 1 and 4 will be used to experiment with high, medium and low environmental changes respectively. As mentioned, in this work the real time is considered, so the values above represents time in seconds between the two consecutive changes. To have an idea, these values represent the following number of fitness evaluations: $0.5 \approx 1000$, $1 \approx 2000$, $4 \approx 9000$. Undoubtedly, these numbers are not constant for all test cases due to different time-complexity of each function and stochastic nature of EA.
    \item \textbf{Building train data set:}
    In this experiment, we explore the effect of using more individuals ($k$-best) of population at each time for training the NN. We change the parameters of NN like batch size, epochs, and number of samples accordingly to have roughly the same timing budget for NN with respect to the overall time in each case. In the case for one individual ($k=1$), we do not limit the overall sample size, so as time increases, the samples aggregate. In other words at every time, NN is trained with all previous times best individuals. The reason is as we consider one individual at each time, the collected samples are a few; hence in order to have a reasonable number of samples we keep the previous samples.
    Conversely, for $k>1$ case, we use a window as the samples aggregation limit window (denoted as $n_w$=5). For this case, we limit the number of samples since otherwise as the time increases, they will exponentially increase. In such case, as we have a constant budget then the time assigned to the EA decreases severely.     

    \item \textbf{Number and mechanism to insert predictions:} In this experiment, number of individuals to be replaced (denoted by $n_p$) with predicted solutions are varied and tested. More number of predicted individuals are created by adding noise to the one predicted value by NN. In this experiment, the added noise is constantly at 10\% of the variable boundaries. In a future study, we do the sensitivity analysis for the noise effect on the results.
    In addition to the number of replaced individuals, two different replacement approaches are also compared. The first one, denoted by NNR, replaces randomly chosen individuals of population  with the predicted solutions. The second one, denoted by NNW, first ranks the individuals of population and then replaces the top worst among them with the predicted solutions.
\end{itemize}

\subsection{Test problems and parameters settings}

We created dynamic environments in two general cases for common functions in literature: Sphere, Rosenbrock and Rastrigin. In the first two experiments, objective function is constant while the constraints change, and for the third and fourth experiments, we define the problem as unconstrained with dynamic objective function. Details of the designed dynamism in each experiment is as Table~\ref{tab:test problem design}.

\begin{figure}[t]
\centerline{\includegraphics[height=1.8in]{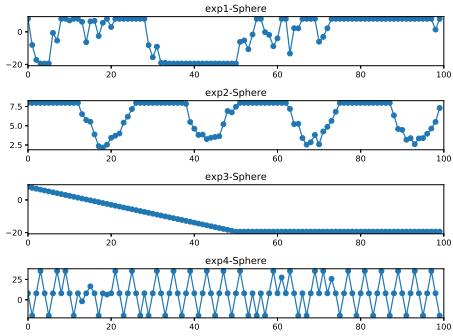}}
    \caption{\scriptsize PCA plot of best\_known positions for each experiment over time}
    \label{fig:pcaPlot}
\end{figure}

\begin{table}
\caption{\scriptsize Designed test problems}
\scalebox{0.9}{
\begin{tabular}{l|l}
\textbf{exp1} &  Uniformly random changes on the boundaries of one linear constraint\\\hline 
\textbf{exp2} &  Patterned sinusoidal changes on the boundaries of one linear constraint\\\hline
\textbf{exp3} &  Linear transformation of the optimum position\\\hline
\textbf{exp4} & \multicolumn{1}{m{8cm}} {Transformation of the optimum position in sinusoidal pattern with random amplitudes}
\end{tabular}
\label{tab:test problem design}
}
\end{table} 
In the first two experiments, the changes are targeted on $b$ values (constraint boundary) of one linear constraint in the the form of $a_ix_i \le b$~\cite{Hasani-ShorehAB19}\footnote{$a_i$ is the coefficient of the variables in the linear constraint}. Figure~\ref{fig:pcaPlot} shows the pattern in which the position of optimum changes in each experiment\footnote{The results belong to best\_known solutions of each time retrieved by executing 100,000 runs of our baseline DE algorithm.}, using principal component analysis (PCA) method to map the thirty dimension to one dimension scale. 

The other parameters are: frequencies of change $\tau$= 0.5, 1, 4; problem dimension=30, runs=30 and the number of changes or times=100.
Parameters of DE are chosen as $NP=20$, $CR=0.2$, $F$ is a random number in $[0.2,0.8]$, and rand/1/bin is the chosen variant of DE~\cite{Ameca-AlducinHB18}. Variable boundaries are limited in $x_i \in [-5,5]$.
Parameters of NN are different based on number of individuals considered to build training data: case $k=1$: epochs=10, $n_w=\infty$ and case $k>1$: epochs=3, $n_w=5$. Also in both cases, we use batch\_size=4, min\_batch=20 and $n_p=3$.
All the experiments were run on a cluster, allocating 1 core (2.4GHz) and 4GB of RAM.
Our code is publicly available on GitHub: $https://github.com/renato145/DENN$.
\subsection{Performance indicators}
\label{subsec:measures}

The applied performance indicators are as follows:

\textbf{Modified offline error ($\text{MOF}$)}
represents the average of the sum of errors in each generation divided by the total generations~\cite{nguyen2012continuous}.

\begin{equation}
MOF= \frac{1}{G_{max}} \sum_{G = 1}^{G_{max}} (|f(\vec{x}^*,t) - f(\vec{x}_{best,G},t)|)
\label{eq:offlineerror}
\end{equation}
Where $G_{max}$ is the maximum generation, $f(\vec{x}^*,t)$ is the global optimum at current time $t$, and $f(\vec{x}_{best,G},t)$ represents the best solution found so far at generation $G$ at current time $t$.
Only feasible solutions are considered to calculate the best errors at every generation. If there were no feasible solution at a particular generation, the worst possible value that a feasible particle can have would be taken.

\textbf{Absolute recovery rate} introduced in~\cite{nguyen2012continuous} is used to analyze the convergence behaviour of algorithms in dynamic environments. This measure infers to how quick an algorithm is to start converging to the global optimum before the next change occurs.

\begin{equation}
ARR= \frac{1}{m}\sum_{i = 1}^{m} (\frac{\sum_{j=1}^{p(i)}|f_{best}(i,j) - f_{best}(i,1)|}{p(i)[f^*(i)-f
_{best}(i,1)]})
\label{eq:offlineerror}
\end{equation}

\textbf{Success rate ($SR$)} calculates in how many times (over all times) each algorithm is successful to reach to $\epsilon$-precision from the global optimum before reaching to the next change. 

\textbf{NN-time} reports the percentage of the time spent to train and use NN per overall optimization time.

\section{EXPERIMENTAL RESULTS}
\label{sec:resultsAnalysis}
In this section, the main findings about designed experiments explained in Section~\ref{sec:ExperimentalSetup} are presented.

\subsection{Frequency changes:}
\begin{figure*}[t]
\centerline{\includegraphics[width=6.5in]{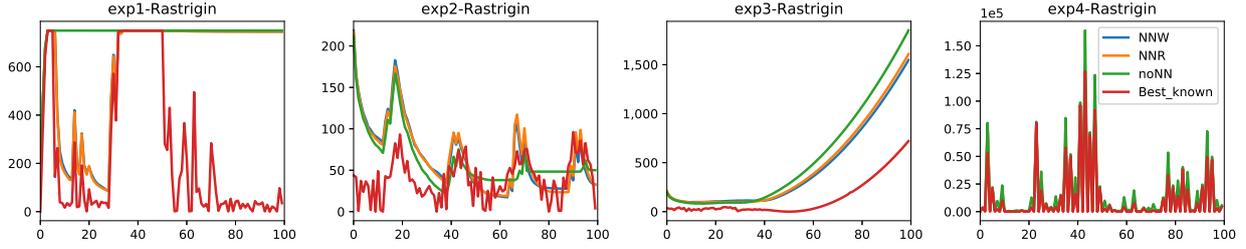}}
    \caption{\scriptsize Fitness values of Rastrigin for $\tau=1$, color-coded with each method over time}
    \label{fig:fitness1}
\end{figure*}
\begin{figure*}
\centerline{\includegraphics[height=3.8in]{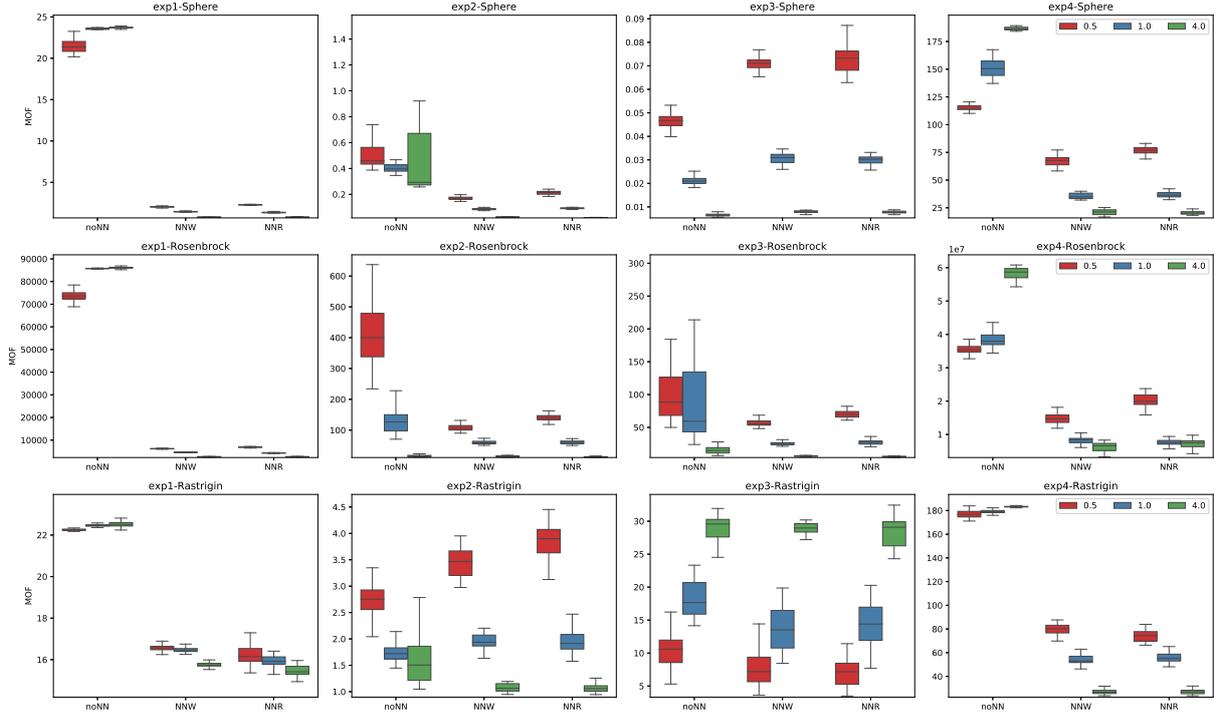}}
    \caption{\scriptsize Distribution of MOF values for each method color-coded with $\tau$ for 30 runs}
    \label{fig:mof}
\end{figure*}
\begin{figure}[t]
\centerline{\includegraphics[width=3.3in,]{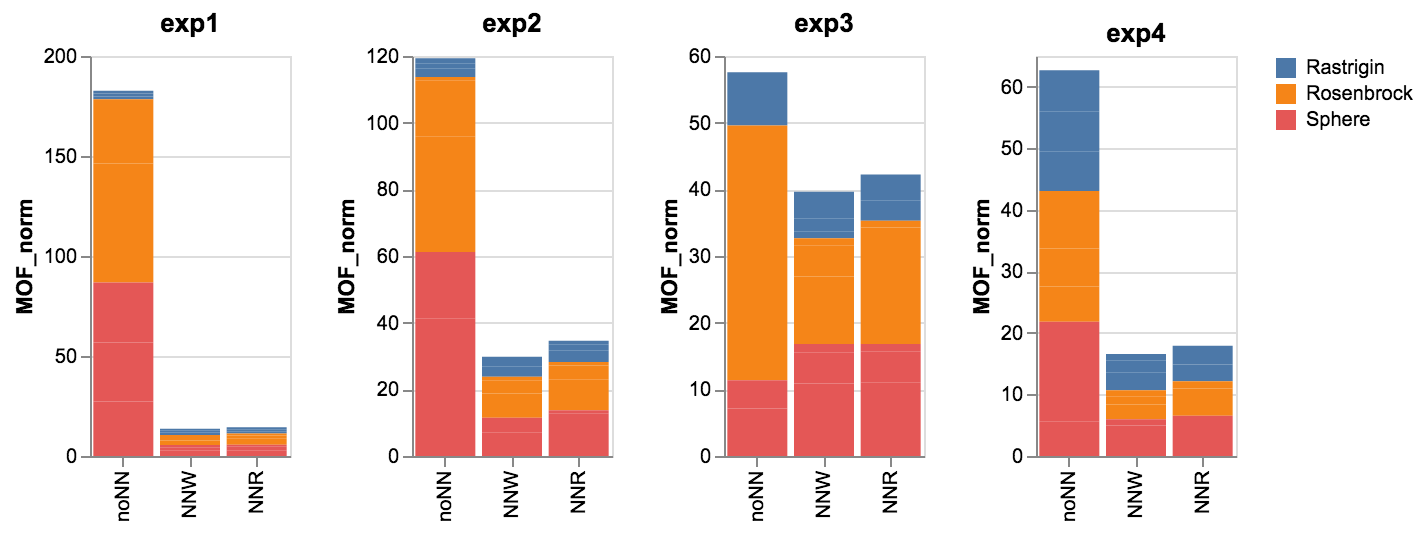}}
    \caption{\scriptsize MOF-norm values considering all frequencies}
    \label{fig:chartAll}
\end{figure}

For most experiments and functions, by increasing $\tau$, $\text{MOF}$ values decrease, presented in Figure~\ref{fig:mof}. 
However, there are some exceptions: for all functions using noNN (exp1 and exp4) and also for Rastrigin function in all methods (exp3).
Looking to PCA plot of the optimum positions in Figure~\ref{fig:pcaPlot} for exp4 (and exp1 for some changes), the optimum alters drastically between two consecutive changes.
As noNN only reevaluates the solutions when a change is detected, they are far away from new optimum and lacking a diversity promotion technique to aid exploring other regions of the search space lead to higher $\text{MOF}$ values.
However, this is only happening in $\tau=4$ as the solutions are more converged in this case compared to the other $\tau$ values.
As for exp1, drastic changes repeat less often, the drop in performance of $\text{MOF}$ value is less severe compared to exp3 (drop in values as $\tau$ increases).
The best fitness values achieved by each method are presented in Figure~\ref{fig:fitness1} over time. From this figure, it is also observable that in most functions and experiments, the best value achieved by NN variants is tracking the optimum more closely.
Figure~\ref{fig:chartAll} illustrates the overall performance of the methods compared to each other color-coded
with different functions considering their performance on all the frequencies. 
Overall comparison of methods is not easily possible with $\text{MOF}$ values as they are not of the same scale. So we use another measure denoted as $\text{MOF\_norm}$ that enables an overall comparison of methods as they represent standard $\text{MOF}$ values (Figure~\ref{fig:chartAll}).
To achieve standard values in each set of function and experiment, the values are divided by the minimum value among all methods. So the method with lowest $\text{MOF}$ value has $\text{MOF\_norm}$ value equal to one and the others are proportionally calculated.

From aforementioned figures (\ref{fig:fitness1},~\ref{fig:mof}, and~\ref{fig:chartAll}) it can be observed that NN variants show their best performance for the experiments where there is a trend in the position changes. Looking to PCA plots (see Figure~\ref{fig:pcaPlot}) for exp3 until the time around 50, we have a linearly decreasing trend and from then it is saturated in variable boundary remaining constant. As the training data for NN depends to previous behaviour of the algorithm, it is unable to self-improve as time passes.   
The NN variants
can obtain better results even in exp1 without a consistent trend. In this experiment, as we consider 5 previous times to train the NN ($n_t=5$), for this $n_t$ there is not a consistent trend observable. 
The better results achieved is partly because the newly generated solutions can increase diversity (as our baseline algorithm lacks a proper diversity mechanism to be activated when a change happens). Thus, even though the change pattern
is not fully consistent, but for
the algorithm without other proper mechanism for reacting to changes still can improve the results.
In addition, this is the reason the difference between MOF values (Figure~\ref{fig:mof}) for this experiment between noNN and NN variants is more significant. Figure~\ref{fig:fitness1} also shows for this frequency, the optimum is not tracked closely for noNN. However, for $\tau$=0.5, as still population has a fair amount of diversity, the optimum is tracked more closely (due to space limitation, we discard other frequencies results).

\begin{figure*}[t]
\begin{subfigure}{.32\textwidth}
\centering
  \includegraphics[height=1.3in]{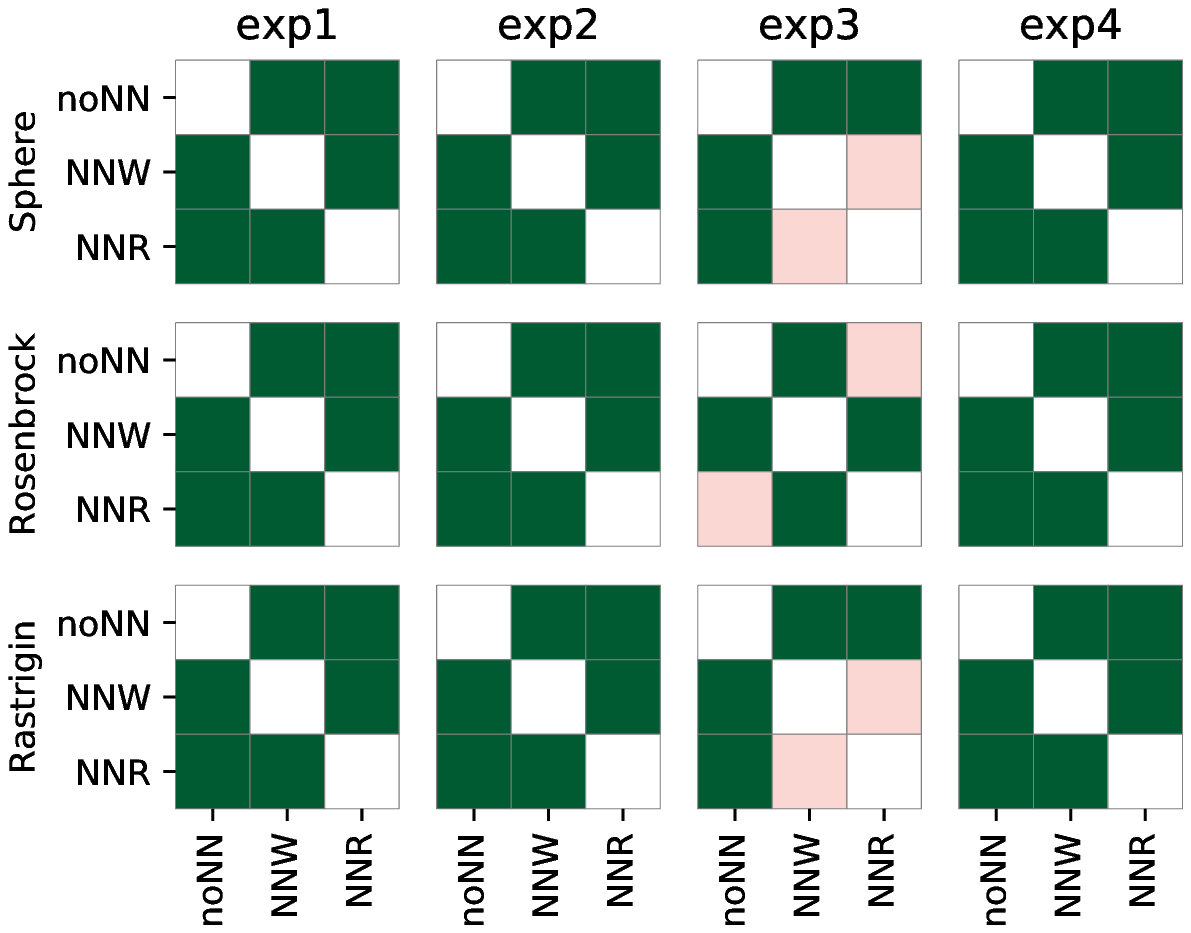}
  \caption{$\tau=0.5$}
  \label{fig:krus0.5}
\end{subfigure}%
\begin{subfigure}{.32\textwidth}
\centering
  \includegraphics[height=1.3in]{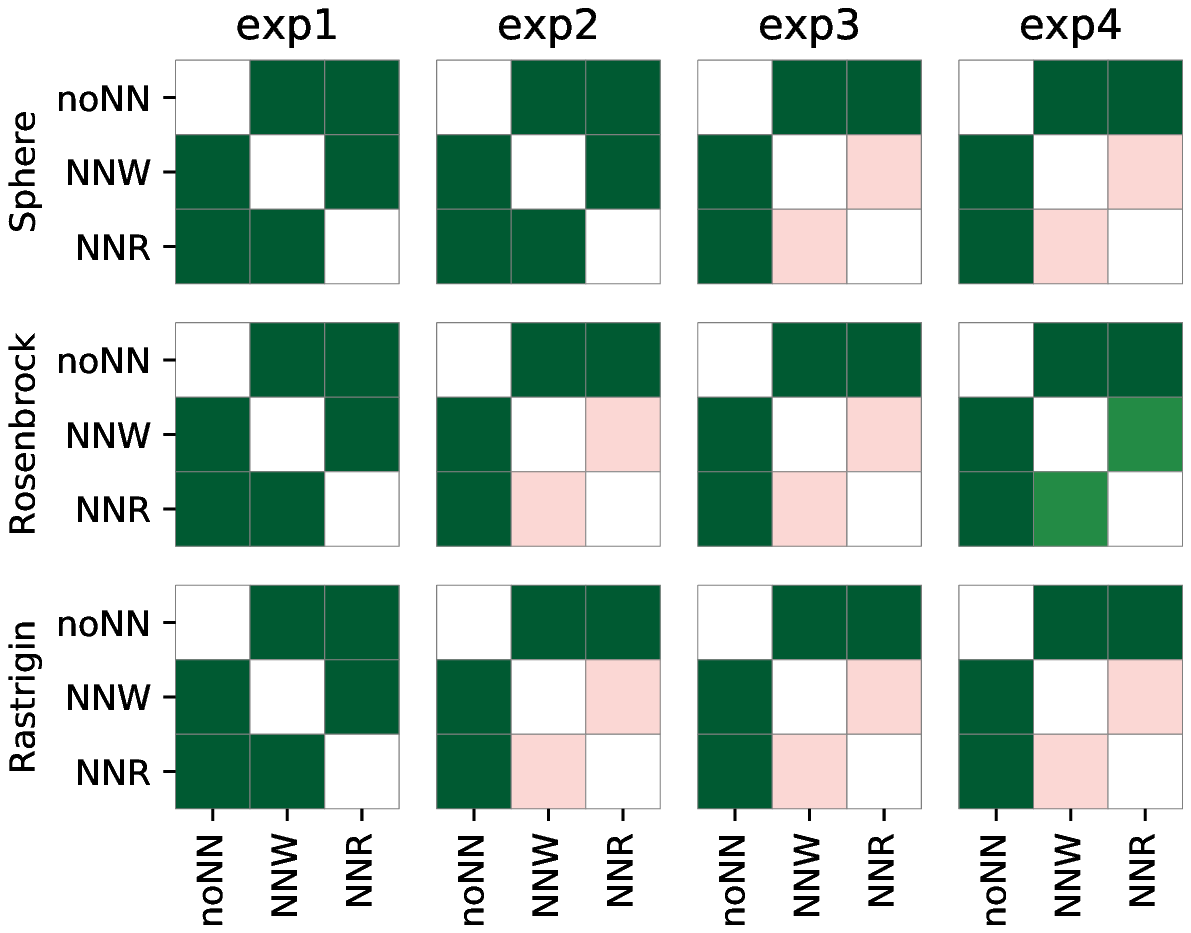} 
  \caption{$\tau=1$}
  \label{fig:krus1}
\end{subfigure}%
\begin{subfigure}{.32\textwidth}
\centering
  \includegraphics[height=1.3in]{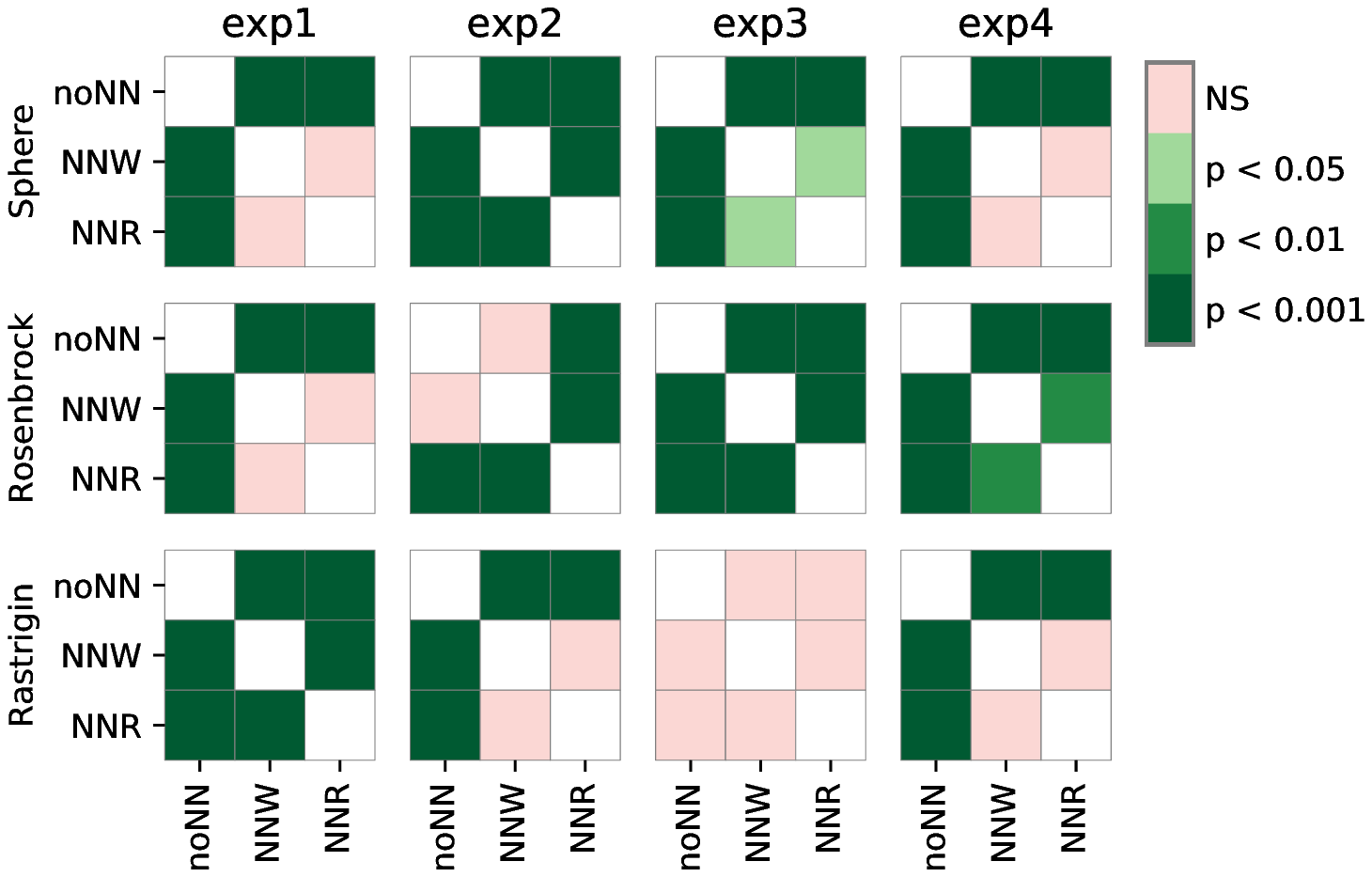} 
  \caption{$\tau=4$}
  \label{fig:krus4}
\end{subfigure}
\caption{\scriptsize Kruskal-Wallis test on MOF values for different frequencies}
\label{fig:krus}
\end{figure*}

To validate the results, the 95\%-confidence Kruskal-Wallis statistical test and the Bonferroni post hoc test, as suggested in~\cite{Derrac20113} are presented. Nonparametric tests were adopted because the samples of runs did not fit to a normal distribution based on the Kolmogorov-Smirnov test.
Figure~\ref{fig:krus} shows a heat-map of the test results on $\text{MOF}$ values. In this figure, as the legend represents, the pink squares show the methods with not-significantly different ($NS$) results, and the squares in the spectrum of the green colors show the significantly different methods with the mentioned $p$-values. 
Results show in most test cases in different frequencies, the methods have significant difference to each
other. However, for higher $\tau$ values (1 and 4) the NN variants show similar behaviour for almost half of the test cases. The reason is as the solutions are converged in high frequencies, there is not significant difference between replacing the worst solutions or select them randomly.

\begin{figure*}[t]
\centerline{\includegraphics[height=3.8in]{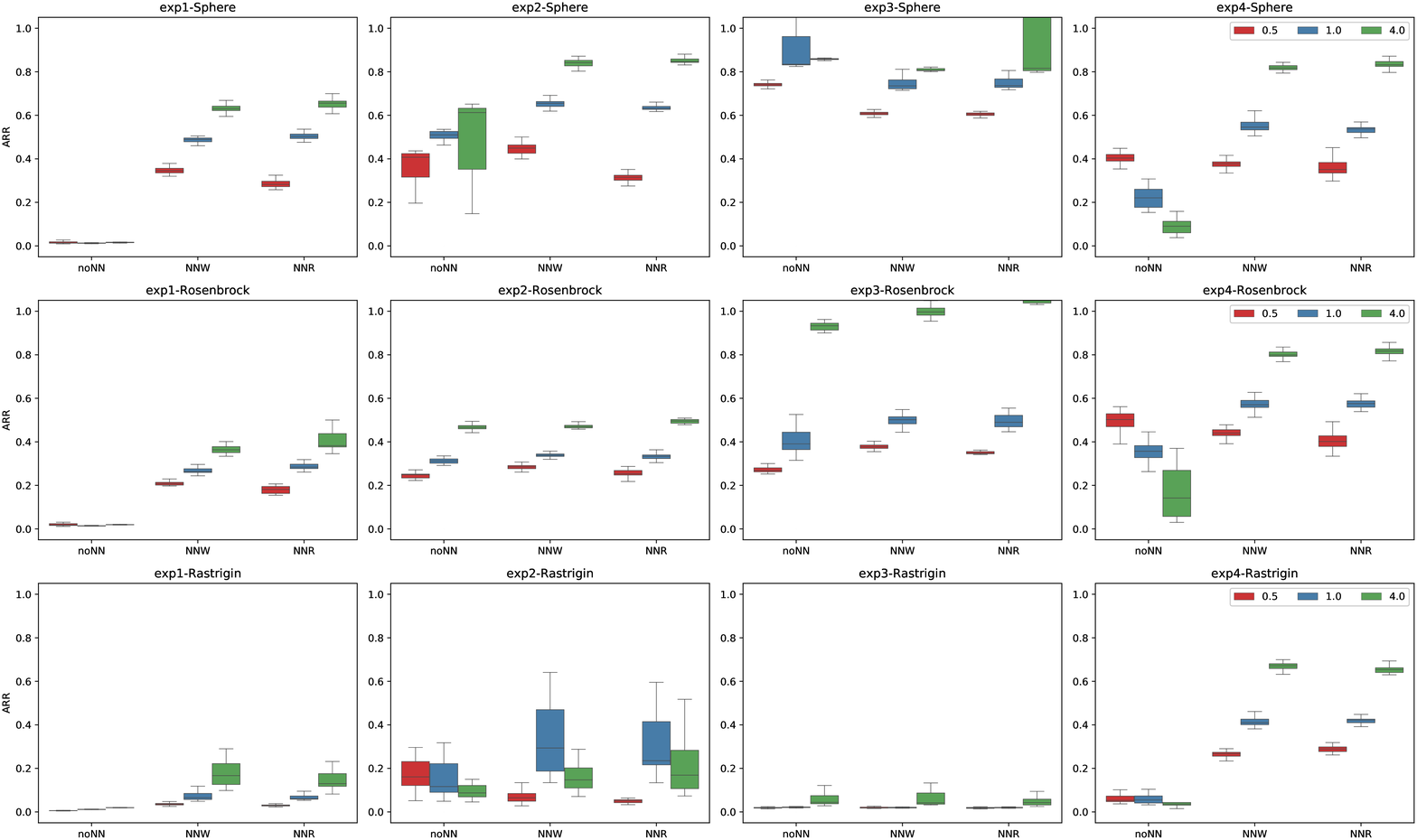}}
    \caption{\scriptsize Distribution of absolute recovery rate (ARR) values color-coded with $\tau$ for 30 runs}
    \label{fig:arr}
\end{figure*}

\begin{figure*}
\centerline{\includegraphics[height=3.8in]{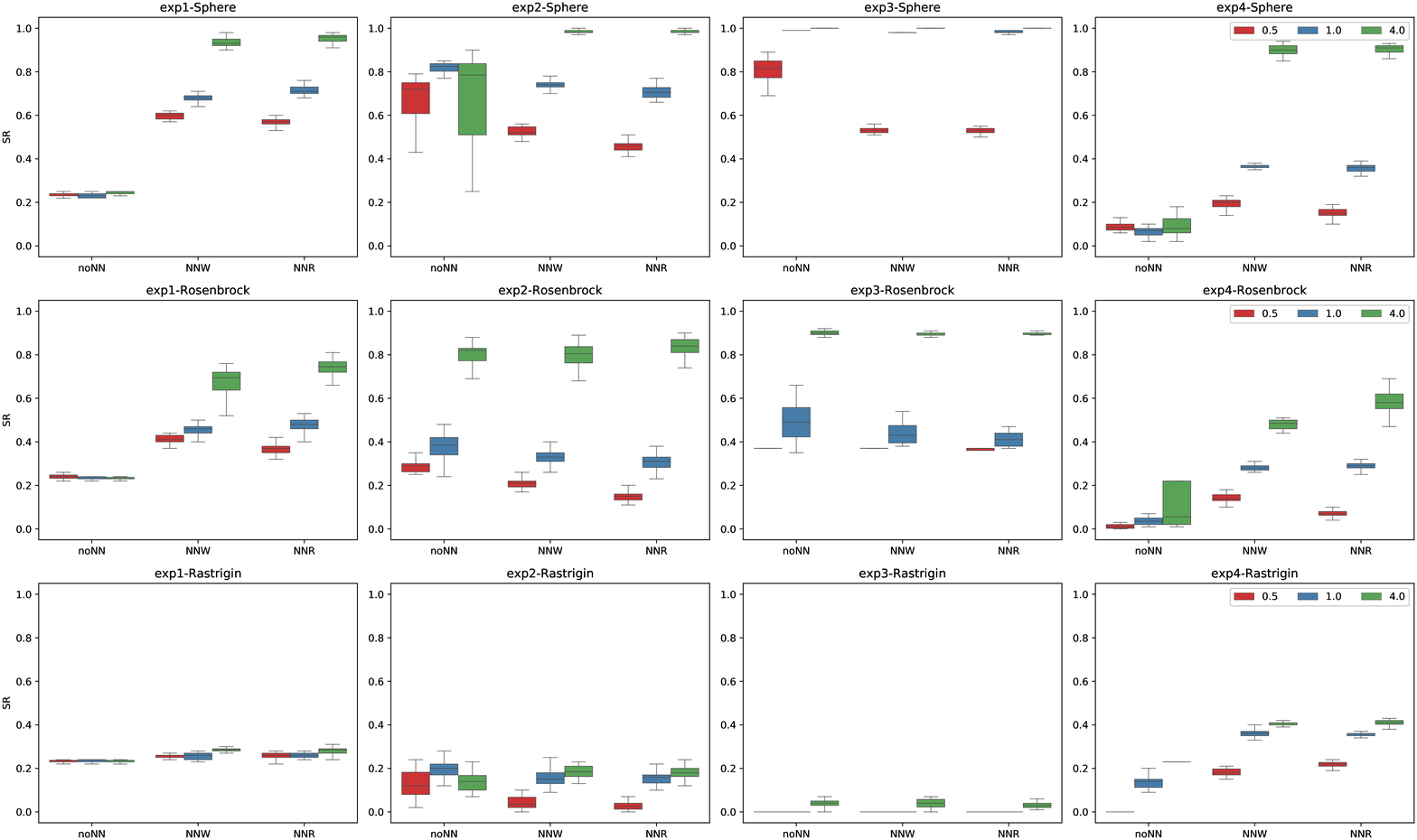}}
    \caption{\scriptsize Distribution of success rate (SR) for 30 runs; number of times algorithms reach to 10\% of the vicinity of optima values per overal times}
    \label{fig:sr}
\end{figure*}

Figures~\ref{fig:arr} and~\ref{fig:sr} show a boxplot of the $\text{ARR}$ and $\text{SR}$ values respectively for different methods and frequencies of change. 
NN variants in most experiments and functions show better $\text{ARR}$ values; meaning they can recover faster after a change. In addition $\text{SR}$ values show better results for NN variants; meaning they can reach to an $\epsilon$-precision (=10\%) of optima for more changes (or times) compared to the method without using prediction. When comparing each method for different frequencies, there is this general trend that better results are achieved as we proceed from frequency 0.5 to 4, as the algorithms have more timing budget to get better results. In addition, NN variants in this frequency, are trained with more precise data as EA has more timing to achieve better solutions.

Table~\ref{tab:nntime} represents the percentages of the amount of time spend for calling NN unit compared to overall optimization time. Regardless of the experiment and function, the results for $\tau=0.5$ show around 20-25\%, $\tau=1$ around 10-12\% and $\tau=4$ around 3\%. This shows when $\tau$ is higher, it is more cheap to use NN in terms of the computational cost. When $\tau$ is low the proportion of the time for doing optimization itself is lower, hence, the samples used to train NN do not represent real optimum or near optimum values and the prediction from NN is not exact in consequence. For this, in most test cases the difference of the performance of NN variants in $\tau=0.5$ and $\tau=4$ is bigger compared to noNN method.

\begin{table*}[t]
\caption{\scriptsize NN-time; time spend for training and using NN in proportion to overall optimization time (mean ± std: for 30 runs)}
\label{tab:nntime}
\centering
\scalebox{0.65}{
\begin{tabular}{|c|c|c|c|c|c|c|c|c|c|c|c|c|c|}
\hline
\multicolumn{2}{|c|}{experiment} & \multicolumn{3}{c|}{exp1} & \multicolumn{3}{c|}{exp2} & \multicolumn{3}{c|}{exp3} & \multicolumn{3}{c|}{exp4} \\ \hline
\multicolumn{2}{|c|}{freq} & 0.5 & 1.0 & 4.0 & 0.5 & 1.0 & 4.0 & 0.5 & 1.0 & 4.0 & 0.5 & 1.0 & 4.0 \\ \hline
\multirow{2}{*}{Sphere} & NNW & 0.24 (±0.00) & 0.11 (±0.00) & 0.02 (±0.00) & 0.25 (±0.00) & 0.10 (±0.00) & 0.03 (±0.00) & 0.24 (±0.00) & 0.12 (±0.00) & 0.03 (±0.00) & 0.19 (±0.00) & 0.10 (±0.00) & 0.03 (±0.00) \\ \cline{2-14} 
 & NNR & 0.21 (±0.02) & 0.12 (±0.03) & 0.03 (±0.00) & 0.22 (±0.03) & 0.12 (±0.01) & 0.03 (±0.00) & 0.24 (±0.01) & 0.11 (±0.00) & 0.03 (±0.00) & 0.19 (±0.00) & 0.12 (±0.00) & 0.02 (±0.00) \\ \hline
\multirow{2}{*}{Rosenbrock} & NNW & 0.23 (±0.00) & 0.12 (±0.00) & 0.02 (±0.00) & 0.22 (±0.00) & 0.11 (±0.00) & 0.02 (±0.00) & 0.23 (±0.00) & 0.11 (±0.00) & 0.03 (±0.00) & 0.22 (±0.00) & 0.11 (±0.00) & 0.02 (±0.00) \\ \cline{2-14} 
 & NNR & 0.20 (±0.02) & 0.12 (±0.02) & 0.03 (±0.00) & 0.21 (±0.02) & 0.12 (±0.01) & 0.03 (±0.00) & 0.24 (±0.01) & 0.11 (±0.00) & 0.03 (±0.00) & 0.19 (±0.01) & 0.12 (±0.00) & 0.03 (±0.00) \\ \hline
\multirow{2}{*}{Rastrigin} & NNW & 0.20 (±0.00) & 0.14 (±0.00) & 0.03 (±0.00) & 0.20 (±0.00) & 0.12 (±0.00) & 0.03 (±0.00) & 0.24 (±0.00) & 0.13 (±0.01) & 0.03 (±0.00) & 0.19 (±0.00) & 0.11 (±0.00) & 0.03 (±0.00) \\ \cline{2-14} 
 & NNR & 0.19 (±0.04) & 0.12 (±0.02) & 0.03 (±0.00) & 0.20 (±0.02) & 0.12 (±0.01) & 0.03 (±0.00) & 0.24 (±0.00) & 0.13 (±0.01) & 0.03 (±0.00) & 0.19 (±0.01) & 0.11 (±0.00) & 0.03 (±0.00) \\ \hline
 \multicolumn{2}{|c|}{Mean values} & 0.21 & 0.12 & 0.03 & 0.22 & 0.12 & 0.03 & 0.24 & 0.12 & 0.03 & 0.2 & 0.11 & 0.03 \\ \hline
\end{tabular}
}
\end{table*}

\subsection{Building train data set:}
\begin{figure}[t]
\centerline{\includegraphics[width=3.3in]{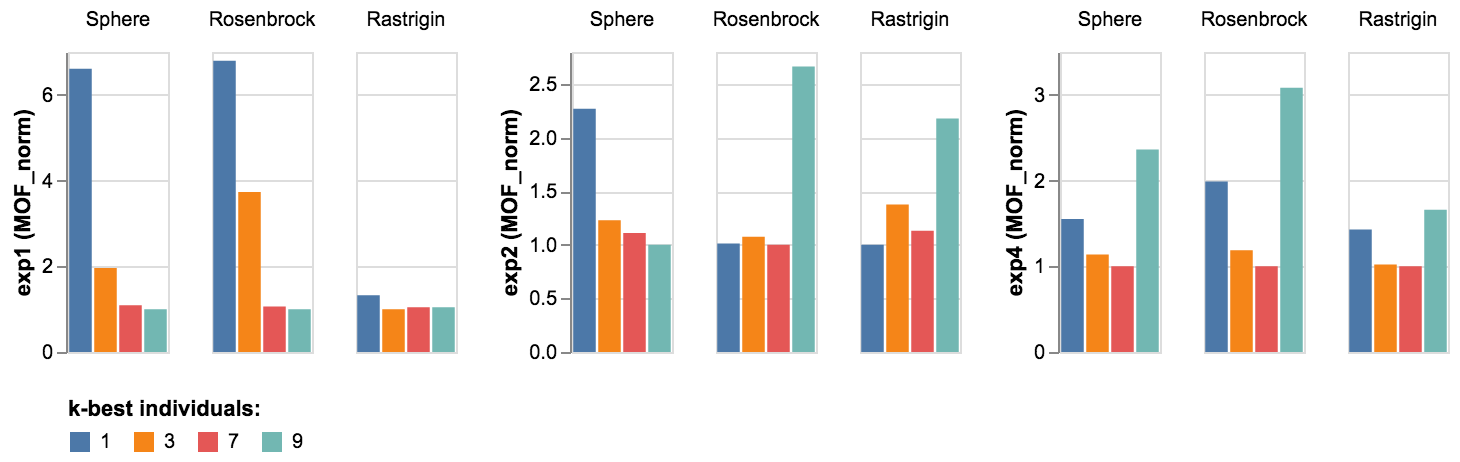}}
    \caption{\scriptsize $k$-best individual selection for building NN samples}
    \label{fig:samplesize}
\end{figure}

We tested 1, 3, 7 and 9 individuals ($k$-best) to be used to train NN.
As Figure~\ref{fig:samplesize} represents, one individual ($k=1$) has not showed good performance based on $\text{MOF\_norm}$ values. The reason is the slow sample collection leads to non-promising MOF values. Due to our min\_batch size (=20), our first prediction is possible at change (time) 26. On the other hand, the results for 9 individuals also degrade. For building our sample data, we take a random combination of solutions for $k$-best solution of each time. Therefore, if the diversity of population is high, the first best solutions are distant from one another and consequently might not represent the change pattern of the optimum correctly. 

Overall, too few or high number of individuals is not a proper choice. So for the rest of the experiments, $k=3$ is chosen to feed NN trainer. Due to space limitation, for this and next experiment, we exclude exp3 to base our conclusions on the experiments where NN performed more promising.

\subsection{Number and mechanism to insert predictions:}
\begin{figure}[t]
\centerline{\includegraphics[width=3.3in]{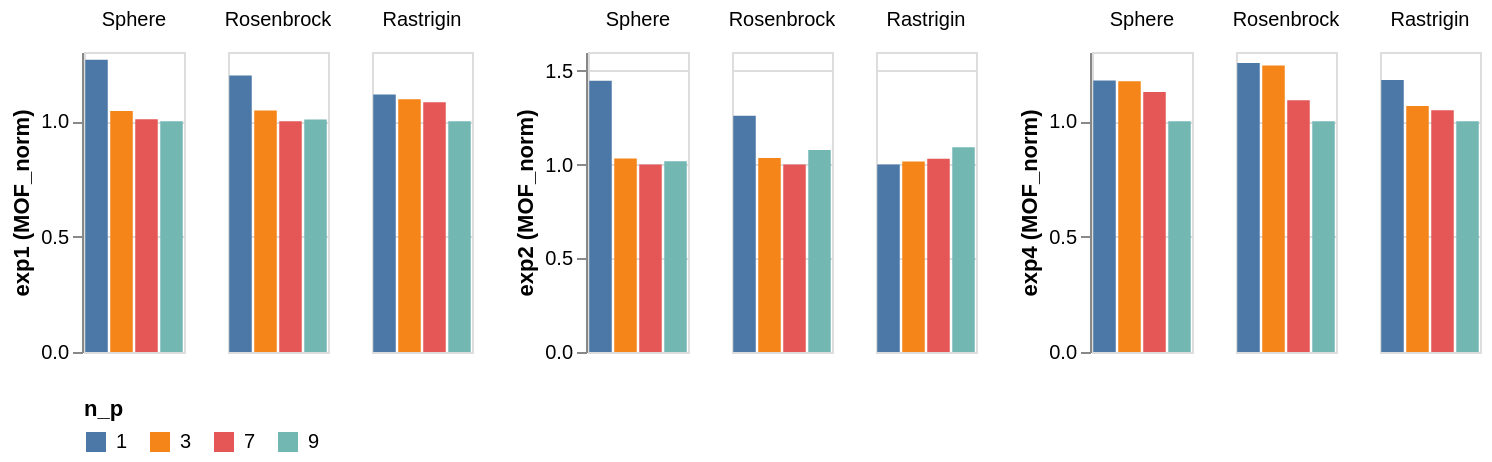}}
    \caption{\scriptsize Number of individuals of population to replace with predicted solutions}
    \label{fig:nnpick}
\end{figure}

When we use a small $n_p$, the effect that NN have in the overall optimization is minor.
On the other hand, using a high $n_p$ will decrease the diversity of the population, given that all the individuals to be included are centered around the same predicted solution with a small added noise (10\% of the variable boundary).
We expect this decrease of diversity to adversely affect the results. However, on our experiments using high $n_p$ values, we do not always observe such a behaviour, as can be seen in exp1 and exp4 (see Figure~\ref{fig:nnpick}).
Also, looking to the pattern of the changes for exp4 (see Figure~\ref{fig:pcaPlot}), the position changes drastically between two alternative times. Since for the rest of the population we only reevaluate the solutions, thus replacing more individuals will help to transfer the population to a new region of the search space.
This is because our baseline algorithm does not promote any diversity, as it only reevaluates the solutions when a change happens. 
Hence, replacing more individuals, particularly for the case with correct predictions, does not have an adverse effect.
However, we believe that in cases where extra mechanisms to promote diversity are considered, the decrease of diversity generated by choosing a high $n_p$ will decrease the overall performance.
In general, there is not significant difference in the results of $\text{MOF}$ values when using $n_p>1$.
In a future work, we plan to explore the effect of the noise added to the predicted solutions on the final performance of the methods. Less noise indicates relying more on the results of the predicted solution. We can have an adaptive noise, that varies based on the results of the prediction error.

Regarding to replace mechanism, based on the results for $\text{MOF}$ values shown in Figure~\ref{fig:chartAll}, we can observe in general that $\text{NNW}$ shows better performance than $\text{NNR}$.
The difference is clearly seen for $\tau$= 0.5, as seen in Figure~\ref{fig:krus0.5}, where there is a significant difference between these two methods for 10 out of 12 test cases.
For larger values of $\tau$ (1 and 4), on the other hand, approximately half of the test cases show significant difference.
The reason for this is the small distance between worst and random picked solutions.
As with a higher $\tau$, all individuals in population are likely to have converged close to the same optimum position.
To conclude, we suggest to insert the predicted solutions by replacing the worst solutions of the population.

\section{CONCLUSION}
\label{sec:conc}

We studied the behaviour of using a NN together with DE for solving DCOPs. 
Considering generated overhead by NN, we observed when the frequency of changes is high, the time spent for NN becomes more noticeable in proportion to overall time. In addition, due to shorter time between changes, the optimization algorithm might not achieve good solutions. In this case, the collected data is not helpful for the prediction or even becomes misleading for the optimization algorithm. In our experiments, for high frequencies of change, NN variants showed their worst results. 

Moreover, for the algorithms integrating with NN enough training data is needed. Hence, for short overall time horizons, this might not be an efficient method as for the first change periods, we need to collect data. Moreover, training a NN with small amounts of data will overfit the NN, making it difficult to generalize and make predictions for new data.
The proposed method to collect more individuals of population from each time to train NN, will lead to make NN ready faster but this is possible when there is lower diversity in population. If the population is diverse, the first best solutions will have higher distance and might not be a good data to train the network.
In general, we observed that diversity has a significant role when applying prediction methods in DCOPs. 
For replacing predicted solutions, we observed when we have diversity among solutions, selection of $n_p$ worst solutions performed better than selecting them randomly.
In general we believe controlling diversity besides prediction methods is essential. To do so, and for a better understanding of the behaviour of the prediction it is suggested to check prediction error and based on that, diversity mechanisms be applied properly together with prediction. We observed in some experiments the lack of diversity lead to poor results, while a basic diversity mechanism could improve results, particularly when predictions are wrong. 
One suggestion for future work is to define an adaptive parameter that considers the prediction error to control to which extent to use diversity mechanisms.

\section*{ACKNOWLEDGEMENTS}
This work has been supported through Australian Research Council (ARC) grants DP160102401 and DP180103232.  

\bibliographystyle{IEEEtran} 

\bibliography{ecai}

\begin{thebibliography}{10}
\providecommand{\url}[1]{#1}
\csname url@samestyle\endcsname
\providecommand{\newblock}{\relax}
\providecommand{\bibinfo}[2]{#2}
\providecommand{\BIBentrySTDinterwordspacing}{\spaceskip=0pt\relax}
\providecommand{\BIBentryALTinterwordstretchfactor}{4}
\providecommand{\BIBentryALTinterwordspacing}{\spaceskip=\fontdimen2\font plus
\BIBentryALTinterwordstretchfactor\fontdimen3\font minus
  \fontdimen4\font\relax}
\providecommand{\BIBforeignlanguage}[2]{{%
\expandafter\ifx\csname l@#1\endcsname\relax
\typeout{** WARNING: IEEEtran.bst: No hyphenation pattern has been}%
\typeout{** loaded for the language `#1'. Using the pattern for}%
\typeout{** the default language instead.}%
\else
\language=\csname l@#1\endcsname
\fi
#2}}
\providecommand{\BIBdecl}{\relax}
\BIBdecl

\bibitem{branke2003designing}
J.~Branke and H.~Schmeck, ``Designing evolutionary algorithms for dynamic
  optimization problems,'' in \emph{Advances in evolutionary computing}.\hskip
  1em plus 0.5em minus 0.4em\relax Springer, 2003, pp. 239--262.

\bibitem{liu2008adaptive}
L.~Liu, E.~M. Zechman, E.~D. Brill, Jr, G.~Mahinthakumar, S.~Ranjithan, and
  J.~Uber, ``Adaptive contamination source identification in water distribution
  systems using an evolutionary algorithm-based dynamic optimization
  procedure,'' in \emph{Water Distribution Systems Analysis Symposium 2006},
  2008, pp. 1--9.

\bibitem{Nguyen20121}
T.~Nguyen, S.~Yang, and J.~Branke, ``Evolutionary dynamic optimization: A
  survey of the state of the art,'' \emph{Swarm and Evolutionary Computation},
  vol.~6, no.~0, pp. 1 -- 24, 2012.

\bibitem{Goh_2009}
C.~K. Goh and K.~C. Tan, ``A competitive-cooperative coevolutionary paradigm
  for dynamic multiobjective optimization,'' \emph{IEEE Transactions on
  Evolutionary Computation}, vol.~13, no.~1, pp. 103--127, Feb 2009.

\bibitem{Bui2005}
L.~T. Bui, H.~A. Abbass, and J.~Branke, ``Multiobjective optimization for
  dynamic environments,'' in \emph{2005 IEEE Congress on Evolutionary
  Computation}, vol.~3, Sept 2005, pp. 2349--2356 Vol. 3.

\bibitem{Richter2013}
H.~Richter, \emph{Evolutionary Computation for Dynamic Optimization
  Problems}.\hskip 1em plus 0.5em minus 0.4em\relax Berlin, Heidelberg:
  Springer Berlin Heidelberg, 2013, ch. Dynamic Fitness Landscape Analysis, pp.
  269--297.

\bibitem{branke2000multi}
J.~Branke, T.~Kau{\ss}ler, C.~Smidt, and H.~Schmeck, ``A multi-population
  approach to dynamic optimization problems,'' in \emph{Evolutionary design and
  manufacture}.\hskip 1em plus 0.5em minus 0.4em\relax Springer, 2000, pp.
  299--307.

\bibitem{Bu_2016}
C.~Bu, W.~Luo, and L.~Yue, ``Continuous dynamic constrained optimization with
  ensemble of locating and tracking feasible regions strategies,'' \emph{IEEE
  Transactions on Evolutionary Computation}, vol.~PP, no.~99, pp. 1--1, 2016.

\bibitem{markov2008evolutionary}
A.~Sim{\~o}es and E.~Costa, ``Evolutionary algorithms for dynamic environments:
  prediction using linear regression and markov chains,'' in
  \emph{International Conference on Parallel Problem Solving from
  Nature}.\hskip 1em plus 0.5em minus 0.4em\relax Springer, 2008, pp. 306--315.

\bibitem{kalman2008tracking}
C.~Rossi, M.~Abderrahim, and J.~C. D{\'\i}az, ``Tracking moving optima using
  kalman-based predictions,'' \emph{Evolutionary computation}, vol.~16, no.~1,
  pp. 1--30, 2008.

\bibitem{autoreg2006dynamic}
I.~Hatzakis and D.~Wallace, ``Dynamic multi-objective optimization with
  evolutionary algorithms: a forward-looking approach,'' in \emph{Proceedings
  of the 8th annual conference on Genetic and evolutionary computation}.\hskip
  1em plus 0.5em minus 0.4em\relax ACM, 2006, pp. 1201--1208.

\bibitem{nonlinearreg2009improving}
A.~Sim{\~o}es and E.~Costa, ``Improving prediction in evolutionary algorithms
  for dynamic environments,'' in \emph{Proceedings of the 11th Annual
  conference on Genetic and evolutionary computation}.\hskip 1em plus 0.5em
  minus 0.4em\relax ACM, 2009, pp. 875--882.

\bibitem{liu2019neural}
X.-F. Liu, Z.-H. Zhan, T.-L. Gu, S.~Kwong, Z.~Lu, H.~B.-L. Duh, and J.~Zhang,
  ``Neural network-based information transfer for dynamic optimization,''
  \emph{IEEE transactions on neural networks and learning systems}, 2019.

\bibitem{meier2018prediction}
A.~Meier and O.~Kramer, ``Prediction with recurrent neural networks in
  evolutionary dynamic optimization,'' in \emph{International Conference on the
  Applications of Evolutionary Computation}.\hskip 1em plus 0.5em minus
  0.4em\relax Springer, 2018, pp. 848--863.

\bibitem{jiang2017transfer}
M.~Jiang, Z.~Huang, L.~Qiu, W.~Huang, and G.~G. Yen, ``Transfer learning-based
  dynamic multiobjective optimization algorithms,'' \emph{IEEE Transactions on
  Evolutionary Computation}, vol.~22, no.~4, pp. 501--514, 2017.

\bibitem{meier2019uncertaint}
A.~Meier and O.~Kramer, ``Predictive uncertainty estimation with temporal
  convolutional networks for dynamic evolutionary optimization,'' in
  \emph{International Conference on Artificial Neural Networks}.\hskip 1em plus
  0.5em minus 0.4em\relax Springer, 2019, pp. 409--421.

\bibitem{ahrari2019new}
A.~Ahrari, S.~Elsayed, R.~Sarker, and D.~Essam, ``A new prediction approach for
  dynamic multiobjective optimization,'' in \emph{2019 IEEE Congress on
  Evolutionary Computation (CEC)}.\hskip 1em plus 0.5em minus 0.4em\relax IEEE,
  2019, pp. 2268--2275.

\bibitem{zhou2013population}
A.~Zhou, Y.~Jin, and Q.~Zhang, ``A population prediction strategy for
  evolutionary dynamic multiobjective optimization,'' \emph{IEEE transactions
  on cybernetics}, vol.~44, no.~1, pp. 40--53, 2013.

\bibitem{simoes2014prediction}
A.~Sim{\~o}es and E.~Costa, ``Prediction in evolutionary algorithms for dynamic
  environments,'' \emph{Soft Computing}, vol.~18, no.~8, pp. 1471--1497, 2014.

\bibitem{bosman2007learning}
P.~A. Bosman and H.~La~Poutre, ``Learning and anticipation in online dynamic
  optimization with evolutionary algorithms: the stochastic case,'' in
  \emph{Proceedings of the 9th annual conference on Genetic and evolutionary
  computation}.\hskip 1em plus 0.5em minus 0.4em\relax ACM, 2007, pp.
  1165--1172.

\bibitem{liu2018neural}
X.-F. Liu, Z.-H. Zhan, and J.~Zhang, ``Neural network for change direction
  prediction in dynamic optimization,'' \emph{IEEE Access}, vol.~6, pp.
  72\,649--72\,662, 2018.

\bibitem{Ameca-AlducinHB18}
M.~Y. Ameca{-}Alducin, M.~Hasani{-}Shoreh, W.~Blaikie, F.~Neumann, and
  E.~Mezura{-}Montes, ``A comparison of constraint handling techniques for
  dynamic constrained optimization problems,'' in \emph{2018 {IEEE} Congress on
  Evolutionary Computation, {CEC} 2018, Rio de Janeiro, Brazil, July 8-13,
  2018}, 2018, pp. 1--8.

\bibitem{deb2000efficient}
K.~Deb, ``An efficient constraint handling method for genetic algorithms,''
  \emph{Computer methods in applied mechanics and engineering}, vol. 186,
  no.~2, pp. 311--338, 2000.

\bibitem{Hasani-ShorehAB19}
M.~Hasani{-}Shoreh, M.~Y. Ameca{-}Alducin, W.~Blaikie, F.~Neumann, and
  M.~Schoenauer, ``On the behaviour of differential evolution for problems with
  dynamic linear constraints,'' in \emph{{IEEE} Congress on Evolutionary
  Computation, {CEC} 2019, Wellington, New Zealand, June 10-13, 2019}, 2019,
  pp. 3045--3052.

\bibitem{nguyen2012continuous}
T.~T. Nguyen and X.~Yao, ``Continuous dynamic constrained optimization—the
  challenges,'' \emph{IEEE Transactions on Evolutionary Computation}, vol.~16,
  no.~6, pp. 769--786, 2012.

\bibitem{Derrac20113}
J.~Derrac, S.~Garc\'ia, D.~Molina, and F.~Herrera, ``A practical tutorial on
  the use of nonparametric statistical tests as a methodology for comparing
  evolutionary and swarm intelligence algorithms,'' \emph{Swarm and
  Evolutionary Computation}, vol.~1, no.~1, pp. 3--18, 2011.

\end{thebibliography}

\end{document}